\documentclass[runningheads]{llncs}
\usepackage[T1]{fontenc}
\usepackage{amsmath}
\usepackage{amsfonts}
\usepackage{amssymb}
\usepackage{adjustbox}
\usepackage{booktabs}
\usepackage{graphicx}
\usepackage{multirow}
\usepackage{xcolor}
\usepackage{comment}
\usepackage{subcaption}
\usepackage{bm}

\usepackage{xspace}

\makeatletter
\DeclareRobustCommand\onedot{\futurelet\@let@token\@onedot}
\def\@onedot{\ifx\@let@token.\else.\null\fi\xspace}

\def\eg{e.g\onedot}

\begin{document}
%
\title{Training-free Zero-shot Composed Image Retrieval via Weighted Modality Fusion and Similarity}
\titlerunning{Training-free ZS-CIR via Weighted Modality Fusion and Similarity}
%

%
%

\author{Ren-Di Wu\orcidID{0009-0007-4564-8790} \and
Yu-Yen Lin\orcidID{{0009-0008-1515-0366}} \and
Huei-Fang Yang\orcidID{0000-0001-8261-6965}}
\authorrunning{R.-D. Wu et al.}
%
\institute{National Sun Yat-sen University, Kaohsiung 804201, Taiwan\\ 
\email{\{b104020009,m124020039\}@student.nsysu.edu.tw}\\
\email{hfyang@mail.cse.nsysu.edu.tw}}


%
\maketitle              
\begin{abstract}
Composed image retrieval (CIR), which formulates the query as a combination of a reference image and modified text, has emerged as a new form of image search due to its enhanced ability to capture user intent.
However, training a CIR model in a supervised manner typically requires labor-intensive collection of (reference image, text modifier, target image) triplets.
While existing zero-shot CIR (ZS-CIR) methods eliminate the need for training on specific downstream datasets, they still require additional pretraining on large-scale image datasets. 
In this paper, we introduce a training-free approach for ZS-CIR.
Our approach, \textbf{Wei}ghted \textbf{Mo}dality fusion and similarity for \textbf{CIR} (WeiMoCIR), operates under the assumption that image and text modalities can be effectively combined using a simple weighted average. 
This allows the query representation to be constructed directly from the reference image and text modifier.
To further enhance retrieval performance, we employ multimodal large language models (MLLMs) to generate image captions for the database images and incorporate these textual captions into the similarity computation by combining them with image information using a weighted average. 
Our approach is simple, easy to implement, and its effectiveness is validated through experiments on the FashionIQ and CIRR datasets. 
Code is available at https://github.com/whats2000/WeiMoCIR.

\end{abstract}
\section{Introduction}
Given a reference image and a textual description, composed image retrieval (CIR) is to find a target image, that visually resembles the reference image and also incorporates modifications specified by the text, from an image database.
Compared to using a single modality (\eg, images or texts alone), which could have limited expressiveness in capturing users' preferences, CIR allows users to better express their preferences with multimodal queries. 
Consequently, recent years have witnessed significant research efforts in CIR~\cite{baldrati-iccv23-SEARLE,baldrati-cvpr22-CLIP4Cir,liu-wacv24-BLIP4Cir,saito-cvpr23-Pic2Word}.

Multimodal queries pose a unique challenge to image retrieval because the retrieval algorithms need to understand what to retain in the visual content and what to modify based on finer details expressed in the natural language.
Previous studies~\cite{baldrati-cvpr22-CLIP4Cir,liu-iccv21-CIRR} have tackled this by training CIR models on large-scale (reference image, text modifier, target image) triplets to learn how to fuse the reference image and the text modifier into a query representation for retrieval.
However, collecting such triplets requires substantial human effort, and the supervised approaches usually exhibit limited performance on unseen data due to being tuned on specific datasets.
Consequently, zero-shot CIR (ZS-CIR), capable of performing CIR without supervised learning on downstream data, has gained attention.
However, existing ZS-CIR methods, such as Pic2Word~\cite{saito-cvpr23-Pic2Word} and SEARLE~\cite{baldrati-iccv23-SEARLE}, still require training on large image datasets to learn mappings for converting images into pseudo-word tokens, which are then combined with textual descriptions for text-to-image retrieval. 
This additional training can be resource-intensive. 

In this paper, we introduce a training-free approach for ZS-CIR. 
Our approach leverages existing vision-language models (VLMs) and multimodal large language models (MLLMs).
Specifically, we use a VLM, such as CLIP~\cite{raford-icml21-CLIP}, to obtain visual features from the reference image and textual features from the text modifier. 
Given that images and texts are aligned within a shared space in the VLMs, we assume that the fused query representation can be obtained through a weighted combination of the visual and textual representations.
During retrieval, unlike previous methods that rely solely on comparing the query feature and database image features for retrieval, our method also incorporates the textual information of the database images for similarity computation.
This is achieved by using an MLLM, such as Gemini~\cite{google-arxiv23-Gemini}, to generate captions for each database image. 
To capture different aspects of the image content, multiple descriptions are generated for each database image. 
By incorporating these generated descriptions, our approach considers both the similarities between the query and visual features of the database images and the similarities between the query and textual features of database images through a weighted average.
We term our approach WeiMoCIR, which stands for \textbf{Wei}ghted \textbf{Mo}dality fusion and similarity for \textbf{CIR}.

Our contributions are summarized as follows:
\begin{itemize}
    \item We introduce a training-free ZS-CIR approach that does not require any resource-intensive training by leveraging pretrained VLMs and MLLMs.
    \item We show that a simple weighted fusion of the image and text modalities is sufficient to generate an effective fused query feature, and incorporating additional textual information further enhances retrieval performance. 
    \item Our method is simple, easy to implement and achieves results comparable to or better than existing methods on FashionIQ~\cite{wu-cvpr21-FashionIQ} and CIRR~\cite{liu-iccv21-CIRR}.
\end{itemize}

\section{Related Work}
\paragraph{Composed Image Retrieval} In CIR, the query comprises two modalities—a reference image and a text modifier—with the goal of retrieving images that satisfy the conditions specified in the query.
Earlier CIR methods~\cite{baldrati-cvpr22-CLIP4Cir,delmas-iclr22-ARTEMIS,vo-cvpr19-TIRG} have focused on developing multimodal fusion techniques to combine the visual features and textual features to obtain the query representation.
For instance, CLIP4CIR~\cite{baldrati-cvpr22-CLIP4Cir} trains a combiner to fuse the information from two modalities.
However, these methods require triplets for specific downstream applications for model training, and collecting such triplets is labor-intensive. 
As a result, recent research has explored ZS-CIR.
Representative methods, like Pic2Word~\cite{saito-cvpr23-Pic2Word} and SEARLE~\cite{baldrati-iccv23-SEARLE}, learn to project images into the textual space using large-scale image datasets. 
With the learned mappings from images to texts, the projected pseudo-word tokens are then combined with the text modifier for text-to-image retrieval without needing to train on downstream datasets.
CIReVL~\cite{karthik-iclr24-CIReVL} further eliminates the need for additional training to learn the image-to-text mappings and achieves training-free ZS-CIR. It utilizes a pretrained VLM~\cite{li-icml22-BLIP,raford-icml21-CLIP} to generate a textual description for the reference image. This description and the text modifier are fed into an LLM~\cite{brown-neurips20-GPT-3,radford-openai19-GPT-2} to produce a textual description of the desired image for text-to-image retrieval. LDRE~\cite{yang-sigir24-LDRE} leverages a pretrained image captioner to generate multiple descriptions for the reference images and employs an LLM to reason based on these descriptions and the modified text. 
This approach demonstrates that multiple image captions can capture the diverse semantics of the composed result, thereby enhancing performance.  

Similar to CIReVL and LDRE, our approach is training-free but differs in two key aspects: (1) our method composes the query representation from both visual and text representation while CIReVL and LDRE translate images to texts; and (2) our method utilizes an MLLM to generate captions for the database images and considers both visual and textual information during retrieval. In contrast, CIReVL and LDRE only consider the visual features of database images.

\paragraph{Vision-Language Models} VLMs~\cite{li-icml22-BLIP,raford-icml21-CLIP}, pretrained on large datasets of image-text pairs, are capable of processing both the visual and textual modalities.
One of the earliest VLMs is CLIP~\cite{raford-icml21-CLIP}. 
Having learned the association between images and texts, CLIP has demonstrated strong generalization capabilities and been applied to various vision-language tasks such as zero-shot classification~\cite{raford-icml21-CLIP}, image generation~\cite{cimino-improve21-CLIP-GLaSS}, and event classification~\cite{li-cvpr22-CLIP-Event}.
Beyond vision-language understanding tasks, BLIP~\cite{li-icml22-BLIP} further extends the capability of the VLMs to generation tasks and can perform tasks such as image captioning~\cite{vinyals-tpami16-NIC} and VQA~\cite{antol-iccv15-Vqa}.

In our work, VLMs are leveraged as feature extractors. Built upon the VLM's joint space where images and texts are aligned, our approach utilizes a weighted approach to fuse the two modalities to achieve ZS-CIR without any training.


\begin{figure}[tb]
    \centering
    \includegraphics[width=1\textwidth]{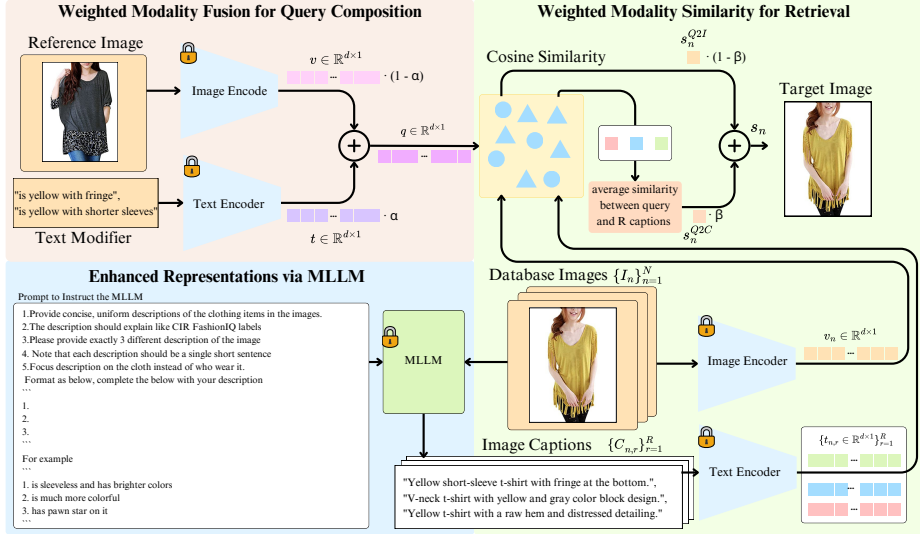}
    \caption{Overview of the proposed WeiMoCIR, a training-free approach for zero-shot composed image retrieval (ZS-CIR). Leveraging pretrained VLMs and MLLMs, our method comprises three modules: weighted modality fusion for query composition, enhanced representations through MLLM-generated image captions, and weighted modality similarity, which integrates both query-to-image and query-to-caption similarities for retrieval.
    }
    \label{fig:method}
\end{figure}

\section{Method}
Our goal is to retrieve a target image from an image database \(\mathcal{D} = \{\bm{I}_n\}_{n=1}^{N}\) that reflects the relevant visual elements of the reference image \( \bm{I} \) along with the modifications specified in the text \(\bm{T}\), all without the need for additional training.
Figure~\ref{fig:method} gives an overview of the proposed WeiMoCIR.
It consists of three modules: the weighted modality fusion module constructs the query representation from the visual and text features; the enhanced representations module improves database image representations with MLLM-generated image captions; and the weighted modality similarity module considers both image and textual information for retrieval. We elaborate on each module in the following.

\paragraph{Weighted Modality Fusion for Query Composition} We utilize a pretrained VLM, such as CLIP or BLIP, which includes a vision encoder and a text encoder as feature extractors. 
For the reference image \(\bm{I}\), the vision encoder \(f_{\theta}\) extracts its visual representation as \(\bm{v} = f_{\theta}(\bm{I}) \in \mathbb{R}^{d \times 1}\), where $d$ is the feature dimension. 
For the text modifier \(\bm{T}\), the text encoder \(f_{\phi}\) extracts a textual representation \(\bm{t} = f_{\phi}(\bm{T}) \in \mathbb{R}^{d \times 1}\).

To find relevant images in the database, an intuitive approach is to use a merger function $\mathcal{M}$ that combines both the reference image and the text and yields a query representation $\bm{q} = \mathcal{M}(\bm{v}, \bm{t})$ for retrieval.
Although the merger function could be complex, we assume that since the images and the texts are aligned in a common space within the VLMs, a simple weighted sum of the visual and textual features can produce a representation with sufficient expressiveness. 
Thus, we obtain the query feature $\bm{q}$ as:
\begin{equation}\label{eqn:query-composition}
    \bm{q} = (1-\alpha) \cdot \bm{v} + \alpha \cdot \bm{t},
\end{equation}
where $\alpha$ controls the weighting between the two modalities.


\paragraph{Enhanced Representations via MLLM-Generated Image Captions} 
While directly comparing the query feature $\bm{q}$ to images in the database by computing query-to-image similarity can yield desired results, we propose leveraging the descriptive power of language to enhance retrieval performance. Language can convey concepts that may be difficult to capture with images alone. 
To capture different aspects of each database image $\bm{I}_n$ in the database, we utilize an MLLM, such as Gemini, to generate a set of $R$ captions, $\{\bm{C}_{n,r}\}_{r=1}^R$, describing the image content from multiple perspectives.

\paragraph{Weighted Modality Similarity for Retrieval} 
Similar to the reference image, for each database image $\bm{I}_n$, the vision encoder extracts its visual representation as \( \bm{v}_n = f_{\theta}(\bm{I}_n) \in \mathbb{R}^{d \times 1} \), and the text encoder is used to obtain its textual representations \( \{\bm{t}_{n,r}\}_{r=1}^R \), where \( \bm{t}_{n,r} = f_{\phi}(\bm{C}_{n,r}) \in \mathbb{R}^{d \times 1} \).

Since the query representation should be close to both the visual representation of the target image and its textual descriptions, we consider both query-to-image and query-to-caption similarities for retrieval. 
Specifically, the query-to-image similarity measures the similarity between the query $\bm{q}$ and the visual representation of a database image $\bm{I}_n$:
\[
s_n^{Q2I} = \mathrm{sim}(\bm{q}, \bm{v}_n),
\]
where \(\mathrm{sim}\) denotes cosine similarity. 
The query-to-caption similarity is the average similarity between the query and the $R$ textual representations of a database image $\bm{I}_n$:
\[
s_n^{Q2C} = \frac{1}{R} \sum_{r=1}^{R} \mathrm{sim}(\bm{q}, \bm{t}_{n,r}).
\]
The final similarity $s_n$ between the query $\bm{q}$ and database image $\bm{I}_n$ is computed as a weighted average of the query-to-image similarity \( s_n^{Q2I} \) and the query-to-caption similarity \( s_n^{Q2C} \):
\begin{equation}\label{eqn:sim-composition}
    s_n = (1 - \beta) \cdot s^{Q2I}_n + \beta \cdot s^{Q2C}_n,
\end{equation}
where $\beta$ controls the influence between two types of similarities.

To find the desired images, the database images are sorted by their similarity scores with respect to the query, and the top \( K \) images are returned.

\section{Experiments}
In this section, we first describe the experimental setup, followed by the experimental results and ablation studies to analyze the effects of the design choices. Finally, qualitative results are presented.

\subsection{Experimental Benchmarks and Protocols}
\paragraph{Datasets} We use FashionIQ~\cite{wu-cvpr21-FashionIQ} and CIRR~\cite{liu-iccv21-CIRR} to validate the effectiveness of the proposed method.
\textbf{FashionIQ}~\cite{wu-cvpr21-FashionIQ} is a dataset focused on fashion apparel, sourced from real-world online data. It contains 30,134 triplet data points from 77,684 images. The dataset is divided into three categories: \textit{Dress}, \textit{Shirt}, and \textit{Toptee}. 
Each triplet is accompanied by two human-generated annotations.
\textbf{CIRR}~\cite{liu-iccv21-CIRR} is a dataset that features a real-world wider range of domains, with more complex descriptive characteristics compared to FashionIQ. The dataset is derived from the NLVR2 \cite{suhr-neurips19-NLVR2} dataset and contains 21,552 images. 

\paragraph{Metrics} 
The evaluation metric is the top K retrieval results (R@K). For FashionIQ, we compute R@K for each of the three categories and their average. 
Since the test set is no longer available for evaluation, we follow previous studies~\cite{baldrati-iccv23-SEARLE,gu-tmlr24-CompoDiff,gu-cvpr24-LinCIR,karthik-iclr24-CIReVL,saito-cvpr23-Pic2Word,zhang-arxiv24-MagicLens} and report results on the validation set. 
For CIRR, we evaluate full-set ranking results using R@1, R@5, R@10, and R@50 metrics. For the subset group, we assess performance using R@1, R@2, and R@3.
We use the validation set to select hyperparameters $\alpha$ and $\beta$ and report the results on the Test1 set.

\subsection{Implementation details}
We use the CLIP model pretrained on LAION-2B English subset of LAION-5B~\cite{scuhmann-neurips22_LAION-5B} as the feature extractor to obtain visual representations from the images and textual representations from the text modifiers and captions. 
For the MLLMs, due to the experiments being conducted at different times, we employed Google Gemini Pro Vision to generate captions for the FashionIQ dataset and Google Gemini 1.5 Flash for the CIRR dataset.
We set $\alpha$ to 0.1 and $\beta$ to 0.8.
All experiments were conducted on a single RTX 3090 GPU with Conda Python 3.8 and PyTorch 2.3.0.

\begin{table}[tb]
    \centering
    \caption{R@K results on FashionIQ. The best scores for each backbone are highlighted in \textbf{bold}.
    The pretrained CLIP is used as the feature extractor.} 
    \label{tab:method_comparison_fashioniq}
    \begin{adjustbox}{max width=\textwidth} 
    \begin{tabular}{@{} @{\extracolsep{\fill} } ll *{8}{c} @{} } 
        \toprule
        &                                              & \multicolumn{2}{c}{Shirt}             & \multicolumn{2}{c}{Dress}             & \multicolumn{2}{c}{Toptee}            & \multicolumn{2}{c}{Average}           \\
        \cmidrule(r){3-4} \cmidrule(r){5-6} \cmidrule(r){7-8} \cmidrule(r){9-10}
        Backbone & Method                              & R@10              & R@50              & R@10              & R@50              & R@10             & R@50               & R@10              & R@50              \\
        \midrule
        \multicolumn{10}{c}{\textit{Methods that need an additional pre-traning step}} \\
        \midrule
        \multirow{5}{*}{ViT L/14}
        & Pic2Word~\cite{saito-cvpr23-Pic2Word}        & 26.20             & 43.60             & 20.00             & 40.20             & 27.90             & 47.40             & 24.70             & 43.70             \\
        & SEARLE-XL-OTI~\cite{baldrati-iccv23-SEARLE}  & 30.37             & 47.49             & 21.57             & 44.47             & 30.90             & 51.76             & 27.61             & 47.90             \\
        & CompoDiff~\cite{gu-tmlr24-CompoDiff}         & \textbf{37.69}    & 49.08             & \textbf{32.24}    & \textbf{46.27}    & \textbf{38.12}    & 50.57             & \textbf{36.02}    & 48.64             \\
        & LinCIR~\cite{gu-cvpr24-LinCIR}               & 29.10             & 46.81             & 20.92             & 42.44             & 28.81             & 50.18             & 26.28             & 46.49             \\
        & MagicLens~\cite{zhang-arxiv24-MagicLens}     & 32.70             & \textbf{53.80}    & 25.50             & 46.10             & 34.00             & \textbf{57.70}    & 30.70             & \textbf{52.50}    \\
        \midrule
        \multirow{3}{*}{ViT H/14}
        & Pic2Word~\cite{saito-cvpr23-Pic2Word}        & \textbf{36.90}    & 55.99             & 28.01             & 51.51             & 40.18             & 62.01             & 35.03             & 56.50             \\
        & SEARLE~\cite{baldrati-iccv23-SEARLE}         & 36.46             & 55.45             & 28.46             & 51.07             & 38.81             & 60.89             & 34.57             & 55.80             \\
        & LinCIR~\cite{gu-cvpr24-LinCIR}               & \textbf{36.90}    & \textbf{57.75}    & \textbf{29.80}    & \textbf{52.11}    & \textbf{42.07}    & \textbf{62.52}    & \textbf{36.26}    & \textbf{57.46}    \\
       \midrule
        \multirow{4}{*}{ViT G/14}
        & Pic2Word~\cite{saito-cvpr23-Pic2Word}        & 33.17             & 50.39             & 25.43             & 47.65             & 35.24             & 57.62             & 31.28             & 51.89             \\
        & SEARLE~\cite{baldrati-iccv23-SEARLE}         & 36.46             & 55.35             & 28.16             & 50.32             & 39.83             & 61.45             & 34.81             & 55.71             \\
        & CompoDiff~\cite{gu-tmlr24-CompoDiff}         & 41.31             & 55.17             & 37.78             & 49.10             & 44.26             & 56.41             & 39.02             & 51.71             \\
        & LinCIR~\cite{gu-cvpr24-LinCIR}               & \textbf{46.76}    & \textbf{65.11}    & \textbf{38.08}    & \textbf{60.88}    & \textbf{50.48}    & \textbf{71.09}    & \textbf{45.11}    & \textbf{65.69}    \\
        \midrule
        \multicolumn{10}{c}{\textit{Training-free methods}} \\
        \midrule
        \multirow{3}{*}{ViT B/32}
        & CIReVL~\cite{karthik-iclr24-CIReVL}         & 28.36             & \textbf{47.84}    & 25.29             & \textbf{46.36}    & 31.21             & 53.85             & 28.29             & 49.35             \\
        & LDRE~\cite{yang-sigir24-LDRE}               & 27.38             & 46.27             & 19.97             & 41.84             & 27.07             & 48.78             & 24.81             & 45.63             \\
        & WeiMoCIR (Ours)                             & \textbf{29.20}    & 47.15             & \textbf{26.23}    & 46.31             & \textbf{34.17}    & \textbf{55.99}    & \textbf{29.86}    & \textbf{49.82}    \\
        \midrule
        \multirow{3}{*}{ViT L/14}
        & CIReVL~\cite{karthik-iclr24-CIReVL}         & 29.49             & 47.40             & 24.79             & 44.76             & 31.36             & 53.65             & 28.55             & 48.57             \\
        & LDRE~\cite{yang-sigir24-LDRE}                   & 31.04             & \textbf{51.22}    & 22.93             & 46.76             & 31.57             & 53.64             & 28.51             & 50.54             \\
        & WeiMoCIR (Ours)                             & \textbf{32.78}    & 48.97             & \textbf{25.88}    & \textbf{47.30}    & \textbf{35.95}    & \textbf{56.71}    & \textbf{31.54}    & \textbf{50.99}    \\
        \midrule  
        \multirow{1}{*}{ViT H/14}
        & WeiMoCIR (Ours)                             & 36.56             & 53.58             & 28.76             & 48.98             & 39.72             & 59.87             & 35.01             & 54.14             \\
        \midrule
        \multirow{3}{*}{ViT G/14}
        & CIReVL~\cite{karthik-iclr24-CIReVL}         & 33.71             & 51.42             & 27.07             & 49.53             & 35.80             & 56.14             & 32.19             & 52.36             \\
        & LDRE~\cite{yang-sigir24-LDRE}               & 35.94             & \textbf{58.58}    & 26.11             & 51.12             & 35.42             & 56.67             & 32.49             & 55.46             \\
        & WeiMoCIR (Ours)                             & \textbf{37.73}    & 56.18             & \textbf{30.99}    & \textbf{52.45}    & \textbf{42.38}    & \textbf{63.23}    & \textbf{37.03}    & \textbf{57.29}    \\
        \bottomrule
    \end{tabular}
    \end{adjustbox}
\end{table}

\begin{table}[tb]
    \centering
    \caption{R@K results on CIRR. The best scores for each backbone are highlighted in \textbf{bold}. The pretrained CLIP is used as the feature extractor.}
    \label{tab:method_comparison_cirr}
    \begin{tabular*}{\textwidth}{@{} @{\extracolsep{\fill}} ll *{7}{c} @{}}
        \toprule
        & & \multicolumn{4}{c}{Recall@K} & \multicolumn{3}{c}{$Recall_{Subset}@K$} \\
        \cmidrule{3-6} \cmidrule{7-9}
        Backbone & Method                             & R@1               & R@5               & R@10              & R@50              & R@1               & R@2               & R@3               \\
        \midrule
        \multicolumn{9}{c}{\textit{Methods that need an additional pretraining step}} \\
        \midrule
        \multirow{5}{*}{ViT L/14}
        & Pic2Word~\cite{saito-cvpr23-Pic2Word}       & 23.90             & 51.70             & 65.30             & 87.80             & 53.76             & 74.46             & 87.08             \\
        & SEARLE-XL-OTI~\cite{baldrati-iccv23-SEARLE} & 24.87             & 52.31             & 66.29             & 88.58             & 53.80             & 74.31             & 86.94             \\
        & CompoDiff~\cite{gu-tmlr24-CompoDiff}        & 18.24             & 53.14             & 70.82             & 90.25             & 57.42             & 77.10             & 87.90             \\
        & LinCIR~\cite{gu-cvpr24-LinCIR}              & 25.04             & 53.25             & 66.68             & -                 & 57.11             & 77.37             & 88.89             \\
        & MagicLens~\cite{zhang-arxiv24-MagicLens}    & \textbf{30.10}    & \textbf{61.70}    & \textbf{74.40}    & \textbf{92.60}    & \textbf{68.10}    & \textbf{84.80}    & \textbf{93.20}    \\
        \midrule
        \multirow{3}{*}{ViT H/14}
        & Pic2Word~\cite{saito-cvpr23-Pic2Word}      & 32.94             & 63.11             & 73.86             & -                 & 62.22             & 81.35             & 91.23             \\
        & SEARLE~\cite{baldrati-iccv23-SEARLE}       & \textbf{34.00}    & \textbf{63.98}    & 75.25             & -                 & \textbf{64.63}    & \textbf{83.21}    & \textbf{92.77}    \\
        & LinCIR~\cite{gu-cvpr24-LinCIR}             & 33.83             & 63.52             & \textbf{75.35}    & -                 & 62.43             & 81.47             & 92.12             \\
        \midrule
        \multirow{4}{*}{ViT G/14}
        & Pic2Word~\cite{saito-cvpr23-Pic2Word}      & 30.41             & 58.12             & 69.23             & -                 & \textbf{68.92}    & \textbf{85.45}    & 93.04             \\
        & SEARLE~\cite{baldrati-iccv23-SEARLE}       & 34.80             & 64.07             & 75.11             & -                 & 68.72             & 84.70             & \textbf{93.23}    \\
        & CompoDiff~\cite{gu-tmlr24-CompoDiff}       & 26.71             & 55.14             & 74.52             & 92.01             & 64.54             & 82.39             & 91.81             \\
        & LinCIR~\cite{gu-cvpr24-LinCIR}             & \textbf{35.25}    & \textbf{64.72}    & \textbf{76.05}    & -                 & 63.35             & 82.22             & 91.98             \\
        \midrule
        \multicolumn{9}{c}{\textit{Training-free methods}} \\
        \midrule
        \multirow{3}{*}{ViT B/32}
        & CIReVL~\cite{karthik-iclr24-CIReVL}        & 23.94             & 52.51             & 66.00             & 86.95             & 60.17             & 80.05             & 90.19             \\
        & LDRE~\cite{yang-sigir24-LDRE}              & 25.69             & 55.13             & 69.04             & 89.90             & \textbf{60.53}    & \textbf{80.65}    & \textbf{90.70}    \\
        & WeiMoCIR (Ours)                            & \textbf{26.31}    & \textbf{57.69}    & \textbf{70.36}    & \textbf{91.01}    & 53.35             & 75.57             & 87.76             \\ 
        \midrule
        \multirow{3}{*}{ViT L/14}
        & CIReVL~\cite{karthik-iclr24-CIReVL}        & 24.55             & 52.31             & 64.92             & 86.34             & 59.54             & 79.88             & 89.69             \\
        & LDRE~\cite{yang-sigir24-LDRE}              & 26.53             & 55.57             & 67.54             & 88.50             & \textbf{60.43}    & \textbf{80.31}    & 89.90             \\
        & WeiMoCIR (Ours)                            & \textbf{30.94}    & \textbf{60.87}    & \textbf{73.08}    & \textbf{91.61}    & 58.55             & 79.06             & \textbf{90.07}    \\
        \midrule
        \multirow{1}{*}{ViT H/14}
        & WeiMoCIR (Ours)                            & 29.11             & 59.76             & 72.34             & 91.18             & 57.23             & 79.08             & 89.76             \\
        \midrule
        \multirow{3}{*}{ViT G/14}
        & CIReVL~\cite{karthik-iclr24-CIReVL}        & 34.65             & 64.29             & 75.06             & 91.66             & 67.95             & 84.87             & 93.21             \\
        & LDRE~\cite{yang-sigir24-LDRE}              & \textbf{36.15}    & \textbf{66.39}    & \textbf{77.25}    & \textbf{93.95}    & \textbf{68.82}    & \textbf{85.66}    & \textbf{93.76}    \\
        & WeiMoCIR (Ours)                            & 31.04             & 60.41             & 72.27             & 90.89             & 58.84             & 78.92             & 89.64             \\
        \bottomrule
    \end{tabular*}
\end{table}

\paragraph{Prompts to Instruct the MLLM to Generate Captions} 
During retrieval, as the captions of the database images are incorporated into the similarity computation, their quality could impact performance. Our experiments show that when the generated captions are more closely aligned with the text modifiers, our method produces more favorable results.
For FashionIQ, the MLLM is instructed to describe the characteristics of the clothes while explicitly avoiding any mention of the person wearing the clothes.
For CIRR, we employ a role-playing instruction to guide the MLLM to provide concise descriptions and exclude any descriptions unrelated to the images. 
For both datasets, we provide a defined output format and a set of example descriptions, and specify that the output should consist of $R$ captions, with $R$ set to 3.


\subsection{Main Experimental Results}
\paragraph{Results on FashionIQ} Table~\ref{tab:method_comparison_fashioniq} shows the comparison of our approach with other ZS-CIR methods, including those requiring additional pretraining and those that are training-free.
It is observed that the zero-shot methods with additional pretraining, such as MagicLens~\cite{zhang-arxiv24-MagicLens} and LinCIR~\cite{gu-cvpr24-LinCIR}, achieve the best performance. This superior performance is primarily due to the additional pretraining that adapts the model to be more suitable for the CIR task.

It is worth noting that the training-free approaches sometimes outperform some methods that involve additional training, demonstrating that favorable results can be achieved by leveraging pretrained models and a well-designed retrieval pipeline. 
Our method consistently surpasses the other training-free approaches across most of the metrics.
Furthermore, our results improve with the use of larger CLIP models on this dataset.
These results validate that leveraging pretrained VLMs and MLLMs, employing a simple weighted fusion of the reference image and text features, and incorporating the textual information into similarity computation are effective strategies for CIR.

\paragraph{Results on CIRR} Table \ref{tab:method_comparison_cirr} shows the results.
Similar to the findings on FashionIQ, the methods requring additional pretraining, particularly those utilizing larger CLIP models, generally have an advantage over the training-free methods. 
However, unlike the results on FashionIQ, our method does not consistently yield better results with larger models.
Hence, as can be seen, our approach yields results comparable to the other training-free methods with smaller CLIP models but underperforms with the larger CLIP ViT-G/14.
This could be due to two reasons:
(1) although larger models generally have stronger generalization capabilities, their performance on specific datasets may not always exceed that of smaller models, and
(2) our approach relies heavily on pretrained VLMs for feature extraction, and the choice of VLM can significantly impact retrieval performance, as confirmed by the ablation study presented next.

\begin{table}[tb]
    \centering
    \caption{Ablation study on the impact of different pretrained VLMs with different backbones on retrieval performance.} 
    \label{tab:ablation-vlms}
    \begin{tabular*}{\linewidth}{@{} @{\extracolsep{\fill}} lcc *{8}{c} @{} } 
        \toprule
        \multicolumn{3}{l}{\textbf{FashionIQ}} & \multicolumn{2}{c}{Shirt} & \multicolumn{2}{c}{Dress} & \multicolumn{2}{c}{Toptee} & \multicolumn{2}{c}{Average} \\
        \cmidrule{4-5} \cmidrule{6-7} \cmidrule{8-9} \cmidrule{10-11}
        Backbone          & $\alpha$ & $\beta$ & R@10           & R@50           & R@10           & R@50           & R@10           & R@50           & R@10           & R@50           \\
        \midrule
        CLIP ViT L/14              & 0.80          & 0.10           & 32.78          & 48.97          & 25.88          & 47.30          & 35.95          & 56.71          & 31.54          & 50.99          \\ 
        CLIP ViT H/14              & 0.80          & 0.10           & 36.56          & 53.58          & 28.76          & 48.98          & 39.72          & 59.87          & 35.01          & 54.14          \\ 
        CLIP ViT G/14              & 0.80          & 0.10           & \textbf{37.73} & \textbf{56.18} & \textbf{30.99} & \textbf{52.45} & \textbf{42.38} & \textbf{63.23} & \textbf{37.03} & \textbf{57.29} \\ 
        BLIP w/ ViT-B              & 0.95          & 0.20           & 23.16          & 37.00          & 20.48          & 38.62          & 25.14          & 47.37          & 22.93          & 41.00          \\
        BLIP w/ ViT-B$^{\dagger}$  & 0.95          & 0.20           & 27.09          & 43.72          & 24.29          & 44.12          & 33.10          & 53.24          & 28.16          & 47.03          \\ 
        BLIP w/ ViT-B$^{\ddagger}$ & 0.95          & 0.20           & 30.52          & 48.33          & 25.33          & 45.46          & 34.63          & 55.89          & 30.16          & 49.90          \\ 
        BLIP w/ ViT-L              & 0.95          & 0.20           & 20.95          & 37.63          & 17.20          & 34.95          & 23.81          & 40.85          & 20.66          & 37.81          \\
        BLIP w/ ViT-L$^{\dagger}$  & 0.95          & 0.20           & 29.15          & 45.73          & 22.86          & 42.89          & 31.11          & 51.56          & 27.70          & 46.72          \\
        BLIP w/ ViT-L$^{\ddagger}$ & 0.95          & 0.20           & 28.95          & 45.83          & 21.22          & 42.44          & 30.55          & 50.33          & 26.91          & 46.20          \\
        \bottomrule
    \end{tabular*}
    \begin{tabular*}{\linewidth}{@{} @{\extracolsep{\fill}} lcc *{7}{c} @{} } 
        \toprule
        \multicolumn{3}{l}{\textbf{CIRR}} & \multicolumn{4}{c}{Recall@K} & \multicolumn{3}{c}{$Recall_{Subset}@K$} \\
        \cmidrule{4-7} \cmidrule{8-10}
        Backbone          & $\alpha$ & $\beta$ & R@1            & R@5            & R@10           & R@50           & R@1            & R@2            & R@3            \\
        \midrule
        CLIP ViT L/14                   & 0.80          & 0.10           & 30.94          & 60.87          & 73.08          & 91.61          & 58.55          & 79.06          & 90.07          \\ 
        CLIP ViT H/14                   & 0.80          & 0.10           & 29.11          & 59.76          & 72.34          & 91.18          & 57.23          & 79.08          & 89.76          \\ 
        CLIP ViT G/14                   & 0.80          & 0.10           & 31.04          & 60.41          & 72.27          & 90.89          & 58.84          & 78.92          & 89.64          \\ 
        BLIP w/ ViT-B              & 0.95          & 0.20           & 25.16          & 52.55          & 64.94          & 86.96          & 56.58          & 77.40          & 88.75          \\ 
        BLIP w/ ViT-B$^{\dagger}$  & 0.95          & 0.20           & 33.37          & 62.63          & 73.30          & 92.19          & 63.98          & 82.46          & 91.81          \\ 
        BLIP w/ ViT-B$^{\ddagger}$ & 0.95          & 0.20           & \textbf{36.51} & \textbf{66.75} & \textbf{77.88} & 93.45          & \textbf{65.06} & \textbf{82.63} & \textbf{92.60} \\ 
        BLIP w/ ViT-L              & 0.95          & 0.20           & 24.46          & 53.04          & 66.70          & 88.99          & 50.92          & 73.78          & 86.65          \\
        BLIP w/ ViT-L$^{\dagger}$  & 0.95          & 0.20           & 30.94          & 61.64          & 73.49          & 92.60          & 57.88          & 78.53          & 90.02          \\
        BLIP w/ ViT-L$^{\ddagger}$ & 0.95          & 0.20           & 32.07          & 63.08          & 75.28          & \textbf{93.49} & 58.63          & 79.13          & 90.53          \\
        \bottomrule
    \end{tabular*}
    \footnotesize{$^{\dagger}$ Finetuned on Flickr30k \quad $^{\ddagger}$ Finetuned on COCO}
\end{table}

\begin{figure}[tb]
    \centering
    \includegraphics[width=1\textwidth]{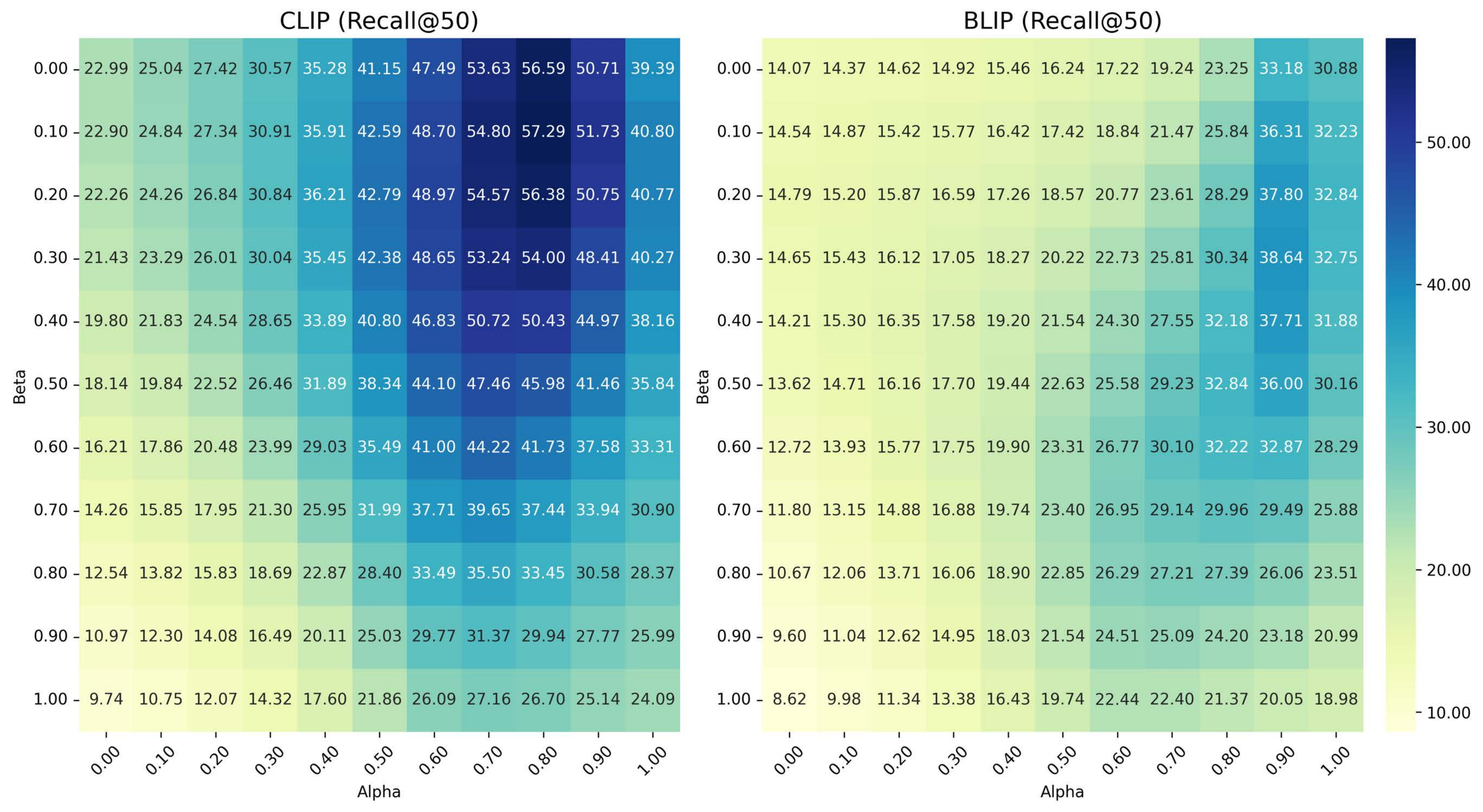}
    \caption{Effects of $\alpha$ and $\beta$ on retrieval performance on FashionIQ. 
    Left: average R@50 using CLIP ViT G/14. Right: average R@50 using BLIP ViT-B.}
    \label{fig:alpha-beta-heat_map}
\end{figure}

\begin{table}[tb]
    \centering
    \caption{Impact of MLLM-generated captions on retrieval performance on the FashionIQ validation set. $\bm{t}_{n,r}$ denotes the $r$the generated caption for the database image $\bm{I}_n$.}
    \label{tab:sample_set_comparison}
    \begin{tabular*}{.9\linewidth}{@{} @{\extracolsep{\fill}} l *{4}{c} @{} } 
        \toprule
         &  \multicolumn{2}{c}{CLIP} & \multicolumn{2}{c}{BLIP} \\
        \cmidrule{2-3} \cmidrule{4-5}
        Captions (\(\bm{t}_{n,r}\)) & R@10 & R@50 & R@10 & R@50 \\
        \midrule
        \(\bm{t}_{n,1}\)                                      & 31.44          & 50.82          & 22.48          & 40.21          \\
        \(\bm{t}_{n,2}\)                                      & 30.77          & 50.18          & 21.33          & 39.48          \\
        \(\bm{t}_{n,3}\)                                      & 30.84          & 49.96          & 22.07          & 38.80          \\
        \(\bm{t}_{n,1} \cup \bm{t}_{n,2}\)                    & 31.84          & 50.94          & 22.79          & 40.83          \\
        \(\bm{t}_{n,2} \cup \bm{t}_{n,3}\)                    & 31.05          & 50.12          & 22.39          & 40.15          \\
        \(\bm{t}_{n,1} \cup \bm{t}_{n,3}\)                    & \textbf{31.77} & 50.98          & \textbf{22.95} & 40.50          \\
        \(\bm{t}_{n,1} \cup \bm{t}_{n,2} \cup \bm{t}_{n,3}\)  & 31.54          & \textbf{50.99} & 22.93          & \textbf{41.00} \\
        \bottomrule
    \end{tabular*}
\end{table}

\subsection{Ablation Study}

\paragraph{Impact of Different Pretrained VLMs with Different Backbones} In Table~\ref{tab:ablation-vlms}, we study the impact of pretrained VLMs on retrieval performance.
The effectiveness of CLIP versus BLIP varies by dataset: CLIP outperforms BLIP on FashionIQ, while both models exhibit comparable performance on CIRR.
When using the BLIP models that have been further finetuned on COCO or Flickr30k, our method achieves significantly better results on both datasets compared to using the BLIP model without additional finetuning.
This improvement is likely due to the high relevance of COCO and Flickr30k images to FashionIQ and CIRR images. 
These findings suggest a strong correlation between the VLMs' pretraining strategies and datasets and their performance on specific downstream datasets.
As our method is training-free, it offers the flexibility to easily employ different VLMs depending on the datasets and applications.

\paragraph{Effects of $\alpha$ and $\beta$ on Performance} 
In Equation~\eqref{eqn:query-composition}, $\alpha$ controls the balance between the visual and textual influence in the query, with higher $\alpha$ values giving more weight to the textual representation.
Similarly, in Equation~\eqref{eqn:sim-composition}, higher $\beta$ values place greater emphasis on the query-to-caption similarity during retrieval.

We analyze the effects of $\alpha$ and $\beta$ on performance using FashionIQ and show the performance heatmaps in Figure~\ref{fig:alpha-beta-heat_map}.
As observed, regardless of the pretrained VLMs used, better retrieval performance is consistently exhibited in the upper-right regions, which correspond to relatively higher $\alpha$ values and lower $\beta$ values.
This suggests that the query representation contains more textual information than visual information.
In the CIR task, where users explicitly specify the desired aspects of the target images through the text modifiers, increasing the influence of textual information in the query representation can lead to more satisfactory retrieval results.

In contrast, the query-to-image similarity is a dominant factor in retrieving the target images, but adding a small amount of information from the query-to-caption similarity can enhance performance. 
We attribute the improvement to the reason that the MLLM-generated captions might provide information not fully captured within the visual representations of the database images.

\begin{figure}[tb]
    \centering
    \begin{subfigure}[b]{\linewidth}
         \centering
         \includegraphics[width=.80\linewidth]{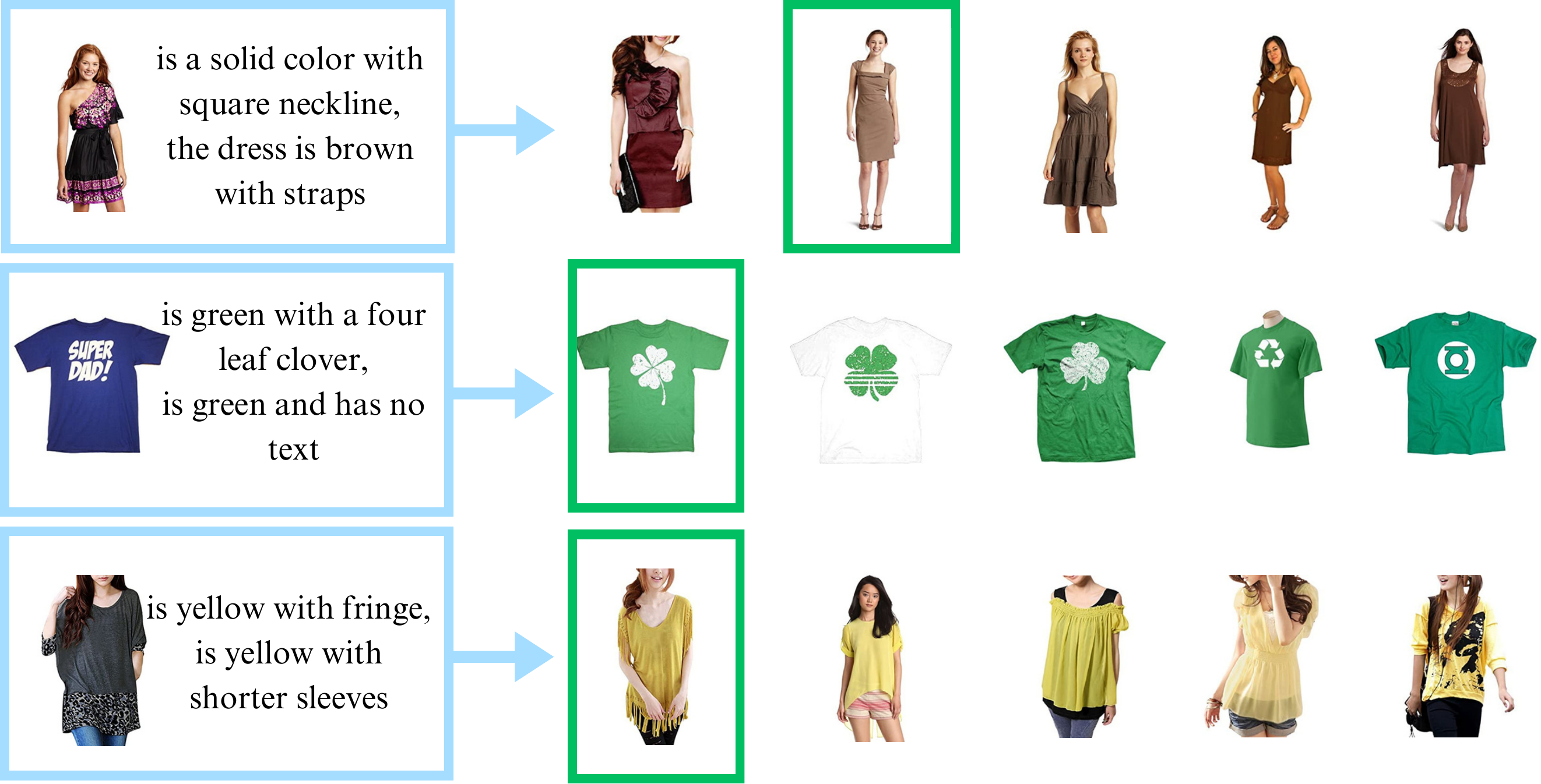}
         \caption{Retrieval results of example queries from FashionIQ.}
         \label{fig:vis-fashioniq}
    \end{subfigure}\vspace{.05in}
    \begin{subfigure}[b]{\linewidth}
         \centering
         \includegraphics[width=.80\linewidth]{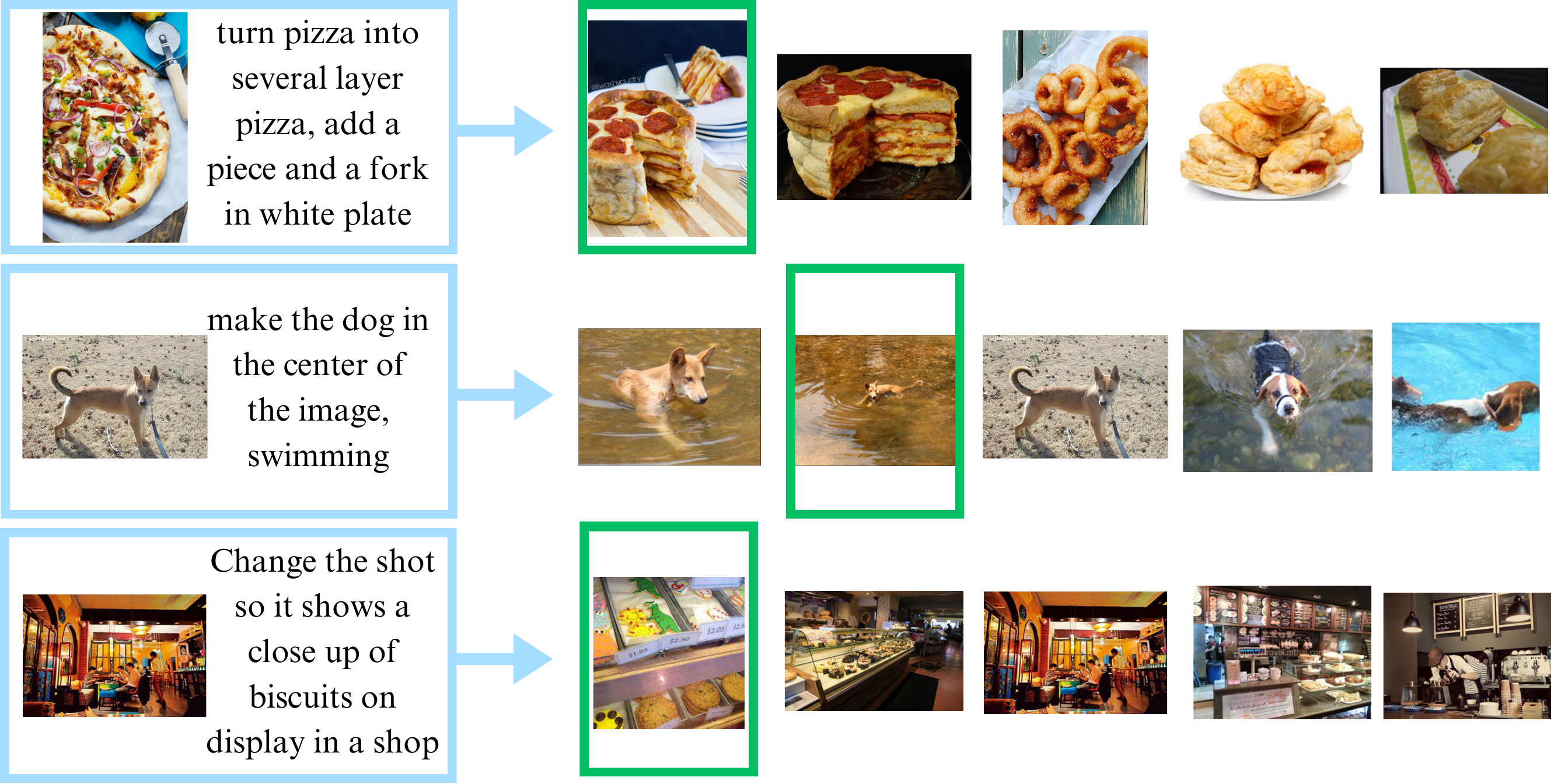}
         \caption{Retrieval results of example queries from CIRR.}
         \label{fig:vis-cirr}
    \end{subfigure}
    \caption{Retrieval results of example queries from FashionIQ and CIRR. Our method with CLIP ViT L/14 successfully retrieves the desired images, highlighted with green boxes, demonstrating its ability to perform CIR for a wide variety of text modifiers.}
    \label{fig:vis-results}
\end{figure}

\paragraph{Effects of MLLM-generated Captions} In our experiments, three captions are generated for each database image. We assess the impact of these MLLM-generated captions on the retrieval performance in Table~\ref{tab:sample_set_comparison}, where $\bm{t}_{n,r}$ denotes the $r$th generated caption for the database image $\bm{I}_n$.

As observed, when only a single image caption is included in the similarity computation, the first caption generated by the MLLM yields better results than the other two, suggesting it captures the image content more effectively. Incorporating multiple captions further enhances retrieval performance, mainly because combining more captions provides a more complete description of the image.
We emphasize that the performance gains from including multiple captions is achieved by guiding MLLMs with well-designed prompts, without any additional human annotations.

\begin{figure}[tb]
    \centering
    \begin{subfigure}[b]{\linewidth}
         \centering
         \includegraphics[width=\linewidth]{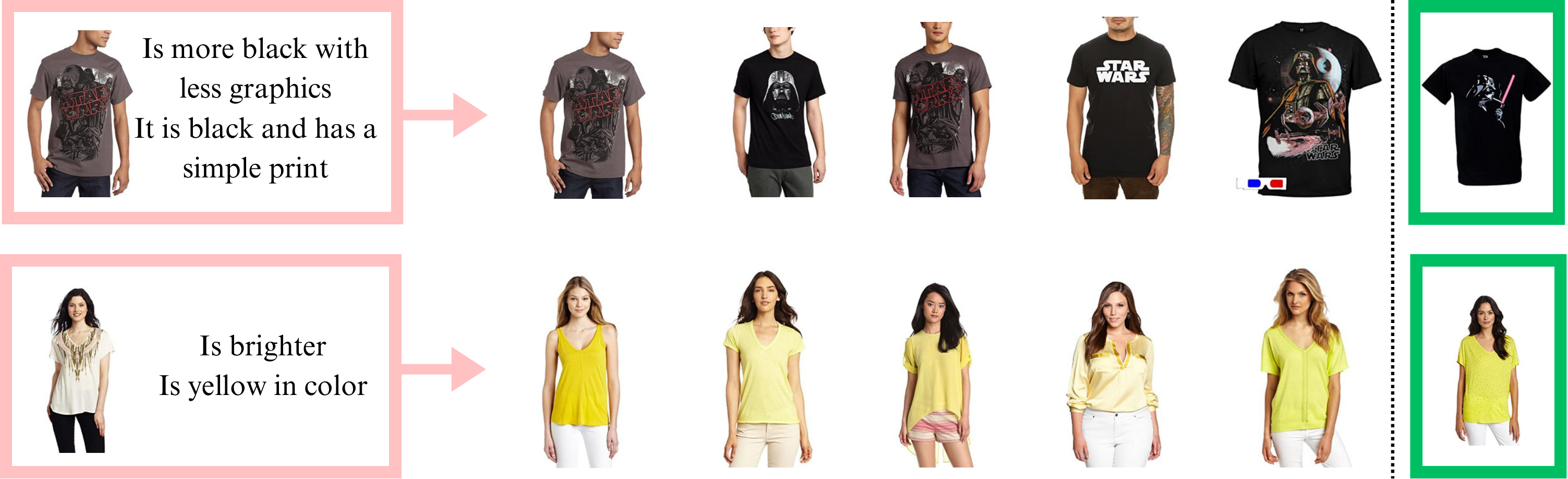}
         \caption{Failure cases of example queries from FashionIQ.}
         \label{fig:vis-fashioniq-fail}
    \end{subfigure}\vspace{.05in}
    \begin{subfigure}[b]{\linewidth}
         \centering
         \includegraphics[width=\linewidth]{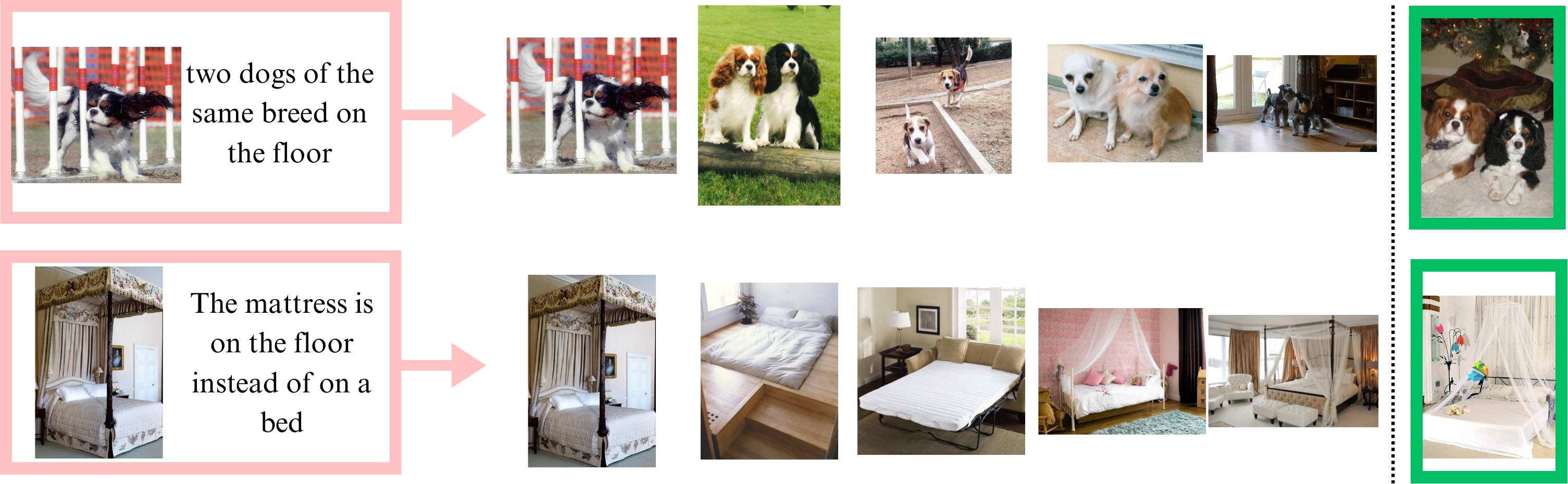}
         \caption{Failure cases of example queries from CIRR.}
         \label{fig:vis-cirr-fail}
    \end{subfigure}
    \caption{Failure cases of our method with CLIP ViT L/14 on example queries from FashionIQ and CIRR. Target images are highlighted with green boxes in the last column. As observed, the failures stem from difficulties in capturing more abstract concepts like `simple print' and recognizing specific `dog breeds'.}
    \label{fig:vis-results-fail}
\end{figure}

\subsection{Qualitative Results}
Figure~\ref{fig:vis-results} shows the CIR results from our method on example queries from FashionIQ and CIRR.
The text modifiers cover a range of modifications, from color and style adjustments in fashion images to complex modifications in scenes, objects, and actions in natural images. 
Our method accurately interprets user intent and retrieves the desired images.

While our method yields promising results, it has certain limitations. As shown in Figure~\ref{fig:vis-results-fail}, although our method can retrieve images that align well with most user requirements, it sometimes struggles to capture the full semantics of complex multimodal queries, such as understanding abstract concepts like `simple print' (1st row of Figure~\ref{fig:vis-fashioniq-fail}), differentiating subtle differences like dog breeds (1st row of Figure~\ref{fig:vis-cirr-fail}), or determining relative object positioning (2nd row of Figure~\ref{fig:vis-cirr-fail}).
These limitations indicate a need for mechanisms to enhance our method's ability to interpret nuanced queries.

\section{Conclusion}

We have presented a training-free approach for ZS-CIR. By leveraging pretrained VLMs and MLLMs, our approach integrates information from different modalities via a weighted average. This results in an easy approach to construct the query representation and allows for the incorporation of textual information into the similarity computation. Our approach performs CIR without any training and hence has the flexibility to use different VLMs and MLLMs depending on the datasets and applications. Experiments and analyses on two datasets demonstrate the effectiveness of our method. Future work will focus on enhancing our method's ability to interpret nuanced queries.

\begin{credits}
\subsubsection{\ackname} This work was supported in part by the National Science and Technology Council, Taiwan under Grant NSTC 113-2640-E-110-002, 112-2221-E-110-047-MY3, and 112-2634-F-006-002.
\end{credits}

%
%
%
\bibliographystyle{splncs04}
\bibliography{refs}

\begin{thebibliography}{10}
\providecommand{\url}[1]{\texttt{#1}}
\providecommand{\urlprefix}{URL }
\providecommand{\doi}[1]{https://doi.org/#1}

\bibitem{antol-iccv15-Vqa}
Antol, S., Agrawal, A., Lu, J., Mitchell, M., Batra, D., Zitnick, C.L., Parikh, D.: Vqa: Visual question answering. In: ICCV. pp. 2425--2433 (2015)

\bibitem{baldrati-iccv23-SEARLE}
Baldrati, A., Agnolucci, L., Bertini, M., Del~Bimbo, A.: Zero-shot composed image retrieval with textual inversion. In: ICCV. pp. 15338--15347 (2023)

\bibitem{baldrati-cvpr22-CLIP4Cir}
Baldrati, A., Bertini, M., Uricchio, T., Del~Bimbo, A.: Effective conditioned and composed image retrieval combining clip-based features. In: CVPR. pp. 21466--21474 (2022)

\bibitem{brown-neurips20-GPT-3}
Brown, T., Mann, B., Ryder, N., Subbiah, M., Kaplan, J.D., Dhariwal, P., Neelakantan, A., Shyam, P., Sastry, G., Askell, A., et~al.: Language models are few-shot learners. NeurIPS pp. 1877--1901 (2020)

\bibitem{cimino-improve21-CLIP-GLaSS}
Cimino, M.G.C.A., Galatolo, F.A., Vaglini, G.: Generating images from caption and vice versa via clip-guided generative latent space search. In: IMPROVE. p. 166–174 (2021)

\bibitem{delmas-iclr22-ARTEMIS}
Delmas, G., Rezende, R.S., Csurka, G., Larlus, D.: Artemis: Attention-based retrieval with text-explicit matching and implicit similarity. In: ICLR (2022)

\bibitem{google-arxiv23-Gemini}
Gemini Team:~Anil, R., Borgeaud, S., Wu, Y., Alayrac, J.B., Yu, J., Soricut, R., Schalkwyk, J., Dai, A.M., Hauth, A., et~al.: Gemini: a family of highly capable multimodal models. arXiv preprint arXiv:2312.11805  (2023)

\bibitem{gu-tmlr24-CompoDiff}
Gu, G., Chun, S., Kim, W., Jun, H., Kang, Y., Yun, S.: {CompoDiff}: Versatile composed image retrieval with latent diffusion. TMLR  (2024)

\bibitem{gu-cvpr24-LinCIR}
Gu, G., Chun, S., Kim, W., Kang, Y., Yun, S.: Language-only training of zero-shot composed image retrieval. In: CVPR. pp. 13225--13234 (2024)

\bibitem{karthik-iclr24-CIReVL}
Karthik, S., Roth, K., Mancini, M., Akata, Z.: Vision-by-language for training-free compositional image retrieval. ICLR  (2024)

\bibitem{li-icml22-BLIP}
Li, J., Li, D., Xiong, C., Hoi, S.: {BLIP}: Bootstrapping language-image pre-training for unified vision-language understanding and generation. In: ICML. pp. 12888--12900 (2022)

\bibitem{li-cvpr22-CLIP-Event}
Li, M., Xu, R., Wang, S., Zhou, L., Lin, X., Zhu, C., Zeng, M., Ji, H., Chang, S.F.: Clip-event: Connecting text and images with event structures. In: CVPR. pp. 16420--16429 (2022)

\bibitem{liu-iccv21-CIRR}
Liu, Z., Rodriguez-Opazo, C., Teney, D., Gould, S.: Image retrieval on real-life images with pre-trained vision-and-language models. In: ICCV. pp. 2125--2134 (2021)

\bibitem{liu-wacv24-BLIP4Cir}
Liu, Z., Sun, W., Hong, Y., Teney, D., Gould, S.: Bi-directional training for composed image retrieval via text prompt learning. In: WACV. pp. 5753--5762 (2024)

\bibitem{raford-icml21-CLIP}
Radford, A., Kim, J.W., Hallacy, C., Ramesh, A., Goh, G., Agarwal, S., Sastry, G., Askell, A., Mishkin, P., Clark, J., Krueger, G., Sutskever, I.: Learning transferable visual models from natural language supervision. In: ICML. pp. 8748--8763 (2021)

\bibitem{radford-openai19-GPT-2}
Radford, A., Wu, J., Child, R., Luan, D., Amodei, D., Sutskever, I., et~al.: Language models are unsupervised multitask learners. OpenAI blog p.~9 (2019)

\bibitem{saito-cvpr23-Pic2Word}
Saito, K., Sohn, K., Zhang, X., Li, C.L., Lee, C.Y., Saenko, K., Pfister, T.: {Pic2Word}: Mapping pictures to words for zero-shot composed image retrieval. In: CVPR. pp. 19305--19314 (2023)

\bibitem{scuhmann-neurips22_LAION-5B}
Schuhmann, C., Beaumont, R., Vencu, R., Gordon, C., Wightman, R., Cherti, M., Coombes, T., Katta, A., Mullis, C., Wortsman, M., et~al.: {LAION-5B}: An open large-scale dataset for training next generation image-text models. In: NeurIPS. pp. 25278--25294 (2022)

\bibitem{suhr-neurips19-NLVR2}
Suhr, A., Zhou, S., Zhang, A., Zhang, I., Bai, H., Artzi, Y.: A corpus for reasoning about natural language grounded in photographs. In: NeurIPS. pp. 6418--6428 (2019)

\bibitem{vinyals-tpami16-NIC}
Vinyals, O., Toshev, A., Bengio, S., Erhan, D.: Show and tell: Lessons learned from the 2015 mscoco image captioning challenge. IEEE TPAMI pp. 652--663 (2016)

\bibitem{vo-cvpr19-TIRG}
Vo, N., Jiang, L., Sun, C., Murphy, K., Li, L.J., Fei-Fei, L., Hays, J.: Composing text and image for image retrieval - an empirical odyssey. In: CVPR. pp. 6439--6448 (2019)

\bibitem{wu-cvpr21-FashionIQ}
Wu, H., Gao, Y., Guo, X., Al-Halah, Z., Rennie, S., Grauman, K., Feris, R.: Fashion iq: A new dataset towards retrieving images by natural language feedback. In: CVPR. pp. 11307--11317 (2021)

\bibitem{yang-sigir24-LDRE}
Yang, Z., Xue, D., Qian, S., Dong, W., Xu, C.: {LDRE}: {LLM}-based divergent reasoning and ensemble for zero-shot composed image retrieval. In: SIGIR. pp. 80--90 (2024)

\bibitem{zhang-arxiv24-MagicLens}
Zhang, K., Luan, Y., Hu, H., Lee, K., Qiao, S., Chen, W., Su, Y., Chang, M.W.: Magiclens: Self-supervised image retrieval with open-ended instructions. arXiv preprint arXiv:2403.19651  (2024)

\end{thebibliography}
\end{document}